\documentclass[letterpaper]{article} 
\usepackage[]{aaai2026}  
\usepackage{times}  
\usepackage{helvet}  
\usepackage{courier}  
\usepackage[hyphens]{url}  
\usepackage{graphicx} 
\usepackage{natbib}  
\usepackage{caption} 
\usepackage{algorithm}

\usepackage{newfloat}
\usepackage{listings}
\DeclareCaptionStyle{ruled}{labelfont=normalfont,labelsep=colon,strut=off} 
\floatstyle{ruled}
\newfloat{listing}{tb}{lst}{}
\floatname{listing}{Listing}
\title{Shedding the Facades, Connecting the Domains: Detecting Shifting  Multimodal Hate Video with Test-Time Adaptation}

\author {
    Jiao Li\textsuperscript{\rm 1, 2},
    Jian Lang\textsuperscript{\rm 1},
    Xikai Tang\textsuperscript{\rm 1},
    Wenzheng Shu\textsuperscript{\rm 3},
    Ting Zhong\textsuperscript{\rm 1}\thanks{Corresponding author.}, 
    Qiang Gao\textsuperscript{\rm 4}, 
    Yong Wang\textsuperscript{\rm 5}, 
    Leiting Chen\textsuperscript{\rm 1, 2},
    Fan Zhou\textsuperscript{\rm 1, 2}
}
\affiliations {
     \textsuperscript{\rm 1}University of Electronic Science and Technology of China\\
     \textsuperscript{\rm 2}Intelligent Digital Media Technology Key Laboratory of Sichuan Province\\
     \textsuperscript{\rm 3}Kuaishou Technology\\
     \textsuperscript{\rm 4}Southwestern University of Finance and Economics\\
     \textsuperscript{\rm 5}Aiwen Technology Co., Ltd.\\
     
     jiao\_li@std.uestc.edu.cn, 
     zhongting@uestc.edu.cn
}

\usepackage{newfloat}
\usepackage{listings}
\usepackage{float}
\usepackage{algorithm}
\usepackage{color}
\usepackage{xcolor} 

\usepackage{bm}
\usepackage{enumitem}
\usepackage{amsmath,amsfonts}
\usepackage{float}
\usepackage{makecell}
\usepackage{xspace}
\usepackage{colortbl}
\usepackage{array}
\usepackage{multirow}
\usepackage{booktabs}
\usepackage{amsthm}
\usepackage{pifont}
\usepackage{subfig}
\usepackage{times}
\usepackage{helvet}
\usepackage{courier}
\usepackage{cleveref}
\usepackage{subcaption}
\usepackage{algpseudocode}
\usepackage{amsmath}    
\usepackage{amsfonts}   
\usepackage{float}

\newcommand{\M}{SCANNER\xspace}
\newcommand{\task}{HVD\xspace}

\usepackage{bibentry}

\begin{document}

\maketitle

\begin{abstract}
Hate Video Detection (\task) is crucial for online ecosystems.
Existing methods assume identical distributions between training (source) and inference (target) data. 
However, hateful content often evolves into irregular and ambiguous forms to evade censorship, resulting in substantial semantic drift and rendering previously trained models ineffective.
Test-Time Adaptation (TTA) offers a solution by adapting models during inference to narrow the cross-domain gap, while conventional TTA methods target mild distribution shifts and struggle with the severe semantic drift in \task.
To tackle these challenges, we propose \textbf{\M}, the first TTA framework tailored for \task. 
Motivated by the insight that, despite the evolving nature of hateful manifestations, their underlying cores remain largely invariant (i.e., targeting is still based on characteristics like gender, race, etc), we leverage these stable cores as a bridge to connect the source and target domains.
Specifically, \M initially reveals the stable cores from the ambiguous layout in evolving hateful content via a principled centroid-guided alignment mechanism.
To alleviate the impact of outlier-like samples that are weakly correlated with centroids during the alignment process, \M enhances the prior by incorporating a sample-level adaptive centroid alignment strategy, promoting more stable adaptation.
Furthermore, to mitigate semantic collapse from overly uniform outputs within clusters, \M introduces an intra-cluster diversity regularization that encourages the cluster-wise semantic richness.
Experiments show that \M outperforms all baselines, with an average gain of 4.69\% in Macro-F1 over the best.
\end{abstract}

\section{Introduction}

The explosive growth of social media has led to a surge in online videos for messaging, which unfortunately foster the creation of hateful multimodal (e.g., audio, text, and visual)  content\footnote{Disclaimer: This paper contains content that may be disturbing to some readers.}. 
In particular, the dissemination of hateful content poses significant social challenges. 
Such hateful content targets specific groups (e.g., race, gender, nationality) to spread discrimination and violence, contributing to social discord and being widely amplified via online platforms~\cite{das2023hatemm, nan2025responsight}. 
Therefore, developing effective methods for Hate Video Detection (\task) has become a pressing and urgent need.

\begin{figure}[t]
    \centering
    \subfloat[Explicit-style]{\includegraphics[height=0.32\columnwidth]{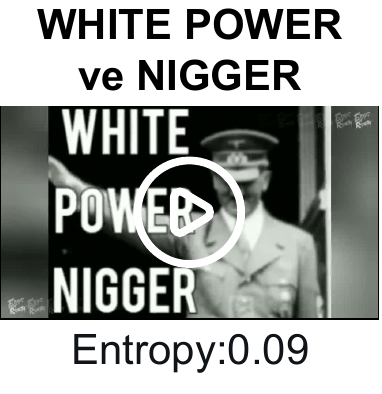}}
    \subfloat[Meme-style]{\includegraphics[height=0.32\columnwidth]{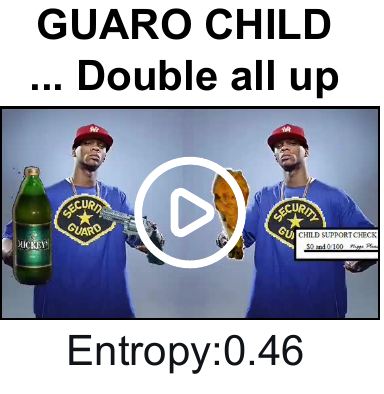}}
    \subfloat[Cartoon-style]{\includegraphics[height=0.32\columnwidth]{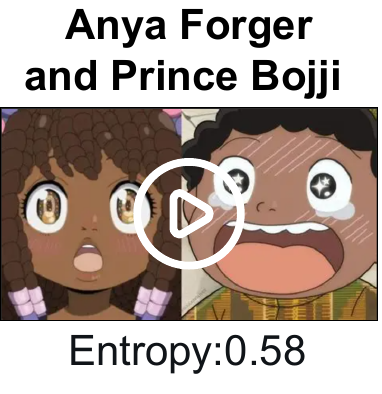}}
    \caption{Examples of evolving hateful manifestations. (a) indicates more explicit forms of hateful content, (b) and (c) represent more implicit and veiled manifestations.} 
    \label{fig:intro}
\end{figure}

Existing works in \task adopt multimodal fusion models for detection~\cite{zhang2023tot, wang2024multihateclip}. 
However, the manifestations of hateful content \textbf{evolve rapidly}~\cite{lang2025biting}, driven by creators’ strategies to evade censorship, attract young audiences, and maximize user engagement.
Specifically, overtly blunt hateful depictions are increasingly replaced by implicit or stylized manifestations, such as garbled or irregular language, meme-style edits, or cartoon imagery (cf.~\Cref{fig:intro}).
As a result, existing methods trained on historical data under the assumption of distributional consistency struggle to detect novel hateful content.
Although MoRE~\cite{lang2025biting} attempts to address this, it requires accessing both source (training) data and target (inference) labels, which is often unrealistic in real-world scenarios.
Due to the extremely sensitive and harmful nature of hateful content, raw training data is often inaccessible and manual annotation is costly and slow, hindering timely detection.
These challenges necessitate a Test-Time Adaptation (TTA) setting in \task, where a model trained on the source domain is expected to adapt to an unseen target domain exhibiting distributional shifts, without access to source data or target labels.

Conventional TTA methods employ self-supervised strategies like entropy minimization or pseudo-label self-training to implicitly bridge cross-domain gaps~\cite{liang2020we, wang2021tent, niu2022efficient, Tsai_2024_CVPR}.
They mainly address mild and regular distributional shifts on originally clean data, such as variations in style or composition.
However, the evolution of hateful content is essentially \textit{an irregular, intent-driven shift in manifestations}, causing substantial yet uncertain semantic drift.
As presented in~\Cref{fig:intro}, while depicting the same hateful topic (i.e., racial discrimination), the left panel shows an explicit manifestation of hateful content, overtly disseminating harmful messages through direct visual symbols and aggressive rhetoric like ``WHITE POWER ve NIGGER''.
In contrast, the center and right panels present ambiguous hateful manifestations, conveyed via humorous or cartoon-style visuals and abstract language like ``GUARO CHILD Double all up ''.
As a result, how to perform effective TTA for \task under severe semantic drift is still an open issue.

To address this, we observe that although the evolving manifestations of hateful content are ceaseless, the hateful cores, such as the targeted demographic categories (e.g., gender, race), remain largely constant. 
As shown in~\Cref{fig:intro}, the targeted groups in all examples predominantly belong to specific racial categories, even though the manifestations of hateful expression vary substantially.
\textit{These invariant hateful cores}, learned and internalized by the source model during training, \textit{can serve as a bridge to connect the source and target domains}.
Drawing upon these observations, a straight thought for tackling the evolving manifestations is to isolate the invariant hateful cores and align target samples toward these cores, reducing sensitivity to surface-level variations while preserving focus on the underlying semantic intent.

Motivated by these insights and the limitations of existing approaches, we design a novel TTA paradigm tailored for \task. 
It aims to guide target samples toward the source domain by eliminating superficial manifestations surrounding the hateful cores and revealing underlying harmful intent.
Initially, to reduce sensitivity to new manifestations, we propose a Centroid-guided AligNment framework, \textbf{CAN}, which strengthens the source-target connection by first clustering target domain samples into hateful centroids  and then encouraging sample alignment toward these invariant cores.
Besides, to prevent the forced alignment of outlier-like samples that are weakly associated with centroids from disrupting gradient optimization (cf.~\Cref{fig:pre_epx2}), we further extend CAN to an advanced Sample-adaptive Centroid AligNment framework, dubbed \textbf{SCAN}.
SCAN dynamically assigns target instances with sample-aware alignment weights based on their affinity to the corresponding centroids, effectively stabilizing the adaptation.
Furthermore, SCAN induces semantic collapse by homogenizing target features around centroids, resulting in overly uniform outputs within clusters (cf.~\Cref{fig:pre_epx1}).
Therefore, we finally upgrade SCAN into \textbf{\M}, a Sample-adaptive Centroid AligNment framework with iNtra-clustER collapse avoidance.
\M incorporates an intra-cluster diversity regularization that promotes cluster-wise semantic diversity of target outputs by encouraging distinct class predictions within clusters.
Our contributions are as follows:
\begin{itemize} [leftmargin=*, topsep=2pt, partopsep=0pt, itemsep=0pt]
\item \M is the pioneering TTA framework specifically designed for \task, aiming to detect unseen and evolving hateful content in videos under drastic and unpredictable semantic drift in manifestations.
\item \M introduces a series of progressively refined alignment frameworks that guide manifestation-evolving hateful samples toward invariant harmful cores, while attenuating the adverse impact of cluster-distant outliers and enhancing intra-cluster semantic diversity.
\item Extensive experiments on three datasets validate the superior adaptation performance of \M, which significantly outperforms all baselines in achieving test-time \task. 
The source codes are available at \url{https://github.com/Jolieresearch/SCANNER}.
\end{itemize}

\section{Related Work}
\label{sec:related-work}
\subsection{Hate Video Detection}
Multimodal malicious content detection is the task of identifying malicious/hateful content expressed through multiple modalities. 
Early efforts focused on memes, leveraging pre-trained vision-language models to analyze their textual and visual components and capture nuanced cross-modal interactions~\cite{cao2023pro, mei2025robust}.
Recently, researchers have turned to more challenging Hate Video Detection (\task), which aims to identify hateful content embedded in visual, textual, and audio modalities within videos. 
They leveraged pre-trained encoders to extract modality features, subsequently utilizing fusion architectures to model their interactions for detection~\cite{das2023hatemm, wang2024multihateclip, koushik2025towards, lang2025biting, hong2025borrowing}.
However, such methods rely on the assumption of distributional consistency, which is violated in real-world settings where hateful content evolves rapidly, leading to significant performance degradation in detection.
Although pivotal work like MoRE~\cite{lang2025biting} begins to address distribution shifts caused by content evolution, it still relies on source data and target labels, resulting in untimely detection in the real world.
In contrast, we present \M for \task under the source-free and target label-free setting, which solely leverages unlabeled target samples to align with the underlying semantics of hateful content by peeling off their superficial manifestations, enabling timely and robust adaptation to evolving hateful content.
\begin{figure*}[ht]
\centering
\includegraphics[width=1.98\columnwidth]{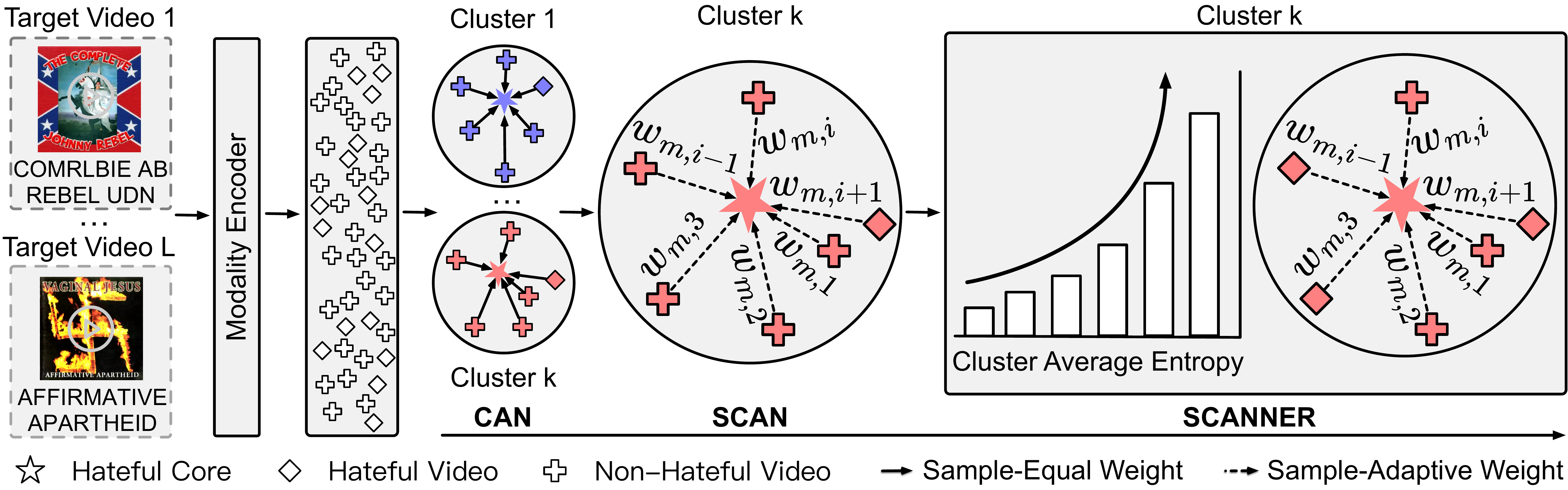}
  \caption{
  An overview of the \M framework, illustrating its progressive evolution from the base framework (CAN) to the final architecture (\M). 
  Video samples are grouped into different clusters (with two example clusters shown; different colors denote different clusters). 
  Solid lines indicate equal weighting for all samples; dashed lines denote similarity-based adaptive weighting, with longer lines implying larger distances from the cluster core and correspondingly smaller weights.
  }
  \label{fig:pipeline}
\end{figure*}
\subsection{Test-Time Adaptation}
Test-Time Adaptation (TTA) aims to bridge the gap between source and target domains through unsupervised fine-tuning the source model using target domain data exhibiting a distribution shift.
Tent updates the channel-wise affine parameters of batch normalization layers by entropy minimization (EM)~\cite{wang2021tent}.
Subsequent research has proposed various EM variants to address challenges such as label shift and continual distribution changes~\cite{wang2022continual, zhou2023ods, gan2023decorate, niu2023towards, liu2023vida, wang2024inter, zhou2025test}.
Recently, some works have delved into the multimodal test-time adaptation~\cite{Cao_2023_ICCV, guo2025bridging, dong2025aeo, guo2025smoothing}.
For example, READ mitigates reliability bias by designing an optimization objective based on EM to narrow the gap between source and target domains~\cite{yang2024test}.
However, most TTA methods are confined to handling standard distribution shifts (e.g., corruption, stylization) for implicitly aligning source and target domains, which are inadequate for addressing irregular and unpredictable manifestation evolution in hateful content.
Consequently, to effectively mitigate the semantic drift in \task, we propose a novel TTA approach, \M.
Our method explicitly leverages the invariant hateful cores (e.g., demographic categories like gender or race) as a bridge to align the source and target domains, effectively narrowing the cross-domain gap.

\section{Methodology}
\subsection{Preliminary}

\noindent \textbf{\task Definition.}
Let $\mathcal{D} =\{D_1, \cdots, D_{N} \}$ denote a \task dataset, where $N$ is the number of videos.
Each video $D_{i}$ comprises visual $v_{i}$ , textual $t_{i}$, and audio $a_{i}$ modalities: $D_{i} = (v_{i}, t_{i}, a_{i})$. 
\task aims to determine whether a video $D_{i}$ contains hateful content (i.e., hateful or non-hateful)~\cite{das2023hatemm, lang2025biting}. 

\noindent \textbf{Test-Time Adaptation Setting.}
A source model $f_{\theta_s}$ is trained on a source domain dataset $\mathcal{D}_s = \{D_i^s\}_{i=1}^{N}$ to learn a function $f_{\theta_s}:(D_i^s) \rightarrow \hat{y}_{i}$, where $D_i^s = (v_{i}^s, t_{i}^s, a_{i}^s, y_i^s)$.
$f_{\theta_s}$ consists of multimodal feature encoders $f_m(\cdot)$, where $m \in \{v, t, a\}$, a transformer-based fusion module $T_s$~\cite{Tsai_2024_CVPR, guo2025bridging} and a classifier $C_s$. 
To mitigate the semantic drift in \task, we adapt the source model $f_{\theta_s}$ on target domain datasets with significant distribution shift caused by the evolving hateful content within videos $\mathcal{D}_t = \{D_i^t\}_{i=1}^{L} = {(v_i^t, t_i^t, a_i^t)}_{i=1}^{L}$ for better detection, under the source-free and target label-free setting.

\noindent \textbf{Overview Framework.}
To better cope with semantic drift in \task, we introduce \M.
\Cref{fig:pipeline} presents an overview of \M framework and illustrates its progressive development: 
starting from the foundational framework CAN (\Cref{CAN}), which encourages target samples to align with invariant hateful cores; 
extending to SCAN (\Cref{SCAN}), which incorporates a sample-level adaptive centroid alignment mechanism into CAN to facilitate sample-level adaptive weighting for more stable alignment; 
and culminating in \M (\Cref{SCANNER}), which further introduces an intra-cluster diversity regularization to prevent semantic collapse during adaptation.

\subsection{CAN: Centroid-Guided Alignment}
\label{CAN}
To adapt the source model for detecting evolving hateful videos at test time, we observe that the underlying hateful cores, such as targeting specific races, genders, or promoting violence, remain largely consistent across the source domain and target domain.
These consistent hateful cores serve as a bridge to narrow the domain gaps, since the source model has already learned to detect such persistent hateful intents during pre-training, but only struggles to generalize to their novel surface manifestations.
In light of these observations, we propose to alleviate the side effect of these novel and ambiguous manifestations during the adaptation by exposing their underlying hateful cores beneath these evolving forms to the source model.
As a result, we initiate a \textbf{C}entroid-guided \textbf{A}lig\textbf{N}ment framework, termed \textbf{CAN}. 
The intuitive objective of CAN is to cluster unlabeled target videos and align them with stable centroids --- representing invariant hateful cores --- to encourage the pre-trained source model to focus on underlying harmful semantics it can recognize with ease, rather than superficial variations.

Concretely, for each modality $m$, where $m \in \{v, t, a\}$, we apply K-Means clustering~\cite{hamerly2003learning, liang2020we} to partition the unlabeled target set $\mathcal{D}_t = \{D_i^t\}_{i=1}^{L}$ into $k$ clusters at the modality level, forming a set of modality-specific centroids ${C}_m = \{c_{m,j}\}_{j=1}^k$:
\begin{equation}
\label{eq:centroid_computation}
c_{m,j} = \frac{1}{|S_{m,j}|} \sum_{D_i^t \in S_{m,j}} f_m(m_i^t), 
\end{equation}
where $S_{m,j}$ denotes the set of the target video $D_i^t$ from $\mathcal{D}_t$ assigned to the $j$-th cluster for modality $m$.
To ensure the stability of the centroids ${C}_m$, we employ a momentum update strategy. 
For the $\tau$-th online adaptation batch, the centroid $c_{m,j}^{(\tau)}$ is updated as follows:
\begin{equation}
\label{eq:momentum_update}
c_{m,j}^{(\tau)} = \gamma \cdot c_{m,j}^{(\tau-1)} + (1-\gamma) \cdot c_{m,j},
\end{equation}
here $c_{m,j}^{(\tau-1)}$ is the centroid from the previous batch and $\gamma \in [0, 1)$ is the momentum coefficient.
Consequently, for each target video $D_i^t$ and modality $m$, we compute its maximum similarity score $s_{m,i}$ to the $k$ centroids, defined as follows:
\begin{equation}
\label{eq:max_similarity_detailed}
s_{m,i} = \max_{j \in \{1, \dots, k\}} \left( \frac{f_m(m_i^t)^{\top} c_{m,j}^{(\tau)}}{\|f_m(m_i^t)\| \cdot \|c_{m,j}^{(\tau)}\|} \right),
\end{equation}
where $c_{m,j}^{(\tau)}$ is the $j$-th centroid for batch $\tau$. 
$\mathcal{L}_{\text{CAN}}$ is a basic clustering loss, which enforces intra-cluster compactness by encouraging each video to be close to their assigned centroids.
$\mathcal{L}_{\text{CAN}}$ can be expressed concisely as the sum of losses over all modalities:
\begin{equation}
\label{eq:bcan_loss_simplified}
\mathcal{L}_{\text{CAN}} = \sum_{m \in \{v, t, a\}} \left( 1 - \frac{1}{B} \sum\nolimits_{i=1}^{B} s_{m,i} \right),
\end{equation}
here $B$ is the number of videos in a mini-batch.
Consequently, minimizing $\mathcal{L}_{\text{CAN}}$ effectively encourages the alignment of target video representations toward their corresponding centroids that capture the invariant hateful cores, thereby leading to tighter and more robust clustering despite the evolving manifestations of hateful videos.

\begin{figure}[t]
    \centering
    \subfloat[From MHY to HMM.]{\includegraphics[height=0.38\columnwidth]{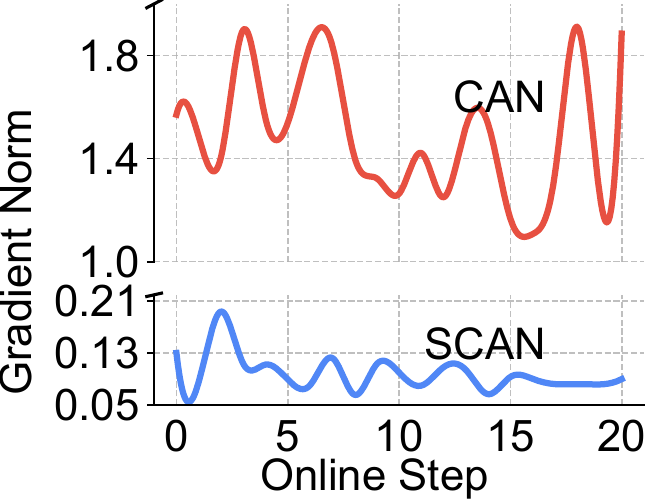}} 
    \subfloat[From MHY to MHB.]{\includegraphics[height=0.38\columnwidth]{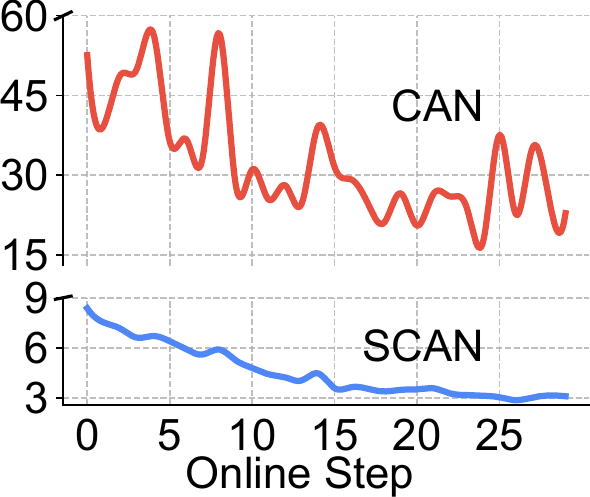}}
    
    \caption{Evolution of gradient norm on CAN and SCAN frameworks during online adaptation.
    Source domain is MHY dataset, target domain are HMM and MHB datasets.
}
    \label{fig:pre_epx2}
\end{figure}

\subsection{SCAN: Sample-Adaptive Alignment}
\label{SCAN}

In the prior CAN framework, aligning each sample equally to respective centroid would forcibly pull some outlier-like samples that are weakly associated with centroids, leading to unstable gradient optimization that hurts the adaptation. 
To verify this hypothesis, we empirically investigate gradient norm behaviour by tracking the $l_2$-norm of all trainable parameter gradients during the optimization of CAN, as shown in~\Cref{fig:pre_epx2}.
The figure clearly shows that the gradient norm of CAN (red line) fluctuates drastically and spikes to abnormally high values, indicating a model collapse.

To get out of this pitfall, we upgrade CAN to a \textbf{S}ample-adaptive \textbf{C}entroid \textbf{A}lig\textbf{N}ment framework (\textbf{SCAN}), which evaluates the similarity between each target sample and its assigned centroid, and subsequently down-weights potential outliers.
Specifically, we advance the loss function $\mathcal{L}_{\text{CAN}}$ into a sample-adaptive version $\mathcal{L}_{\text{SCAN}}$ by introducing a sample-adaptive weight $w_{m,i}$ for each target video $D_i^t$:
\begin{equation}
\label{eq:adaptive_weight}
w_{m,i} = \text{softmax} (\beta \cdot s_{m,i}) = \frac{\exp(\beta \cdot s_{m,i})}{\sum_{j=1}^{B} \exp(\beta \cdot s_{m,j})},
\end{equation}
where $\beta$ is a temperature parameter controlling the smoothness, and $s_{m,i}$ is the maximum similarity score.
Finally, we formulate the sample-adaptive alignment loss:
\begin{equation}
\mathcal{L}_{\text{SCAN}} = \sum_{m \in \{v,t,a\}} \left( 1 - \sum_{i=1}^{B} w_{m,i} \cdot s_{m,i} \right).
\end{equation}
$\mathcal{L}_{\text{SCAN}}$ ensures that the source model prioritizes the alignment of centroid-close samples and suppresses the adverse effects of outliers
Empirical results in~\Cref{fig:pre_epx2} show that, comparing to CAN, SCAN (blue line) effectively stabilizes gradient norms of alignment within a reasonable range.

\begin{figure}[t]
    \centering
    \subfloat[From MHY to HMM.]{\includegraphics[height=0.37\columnwidth]{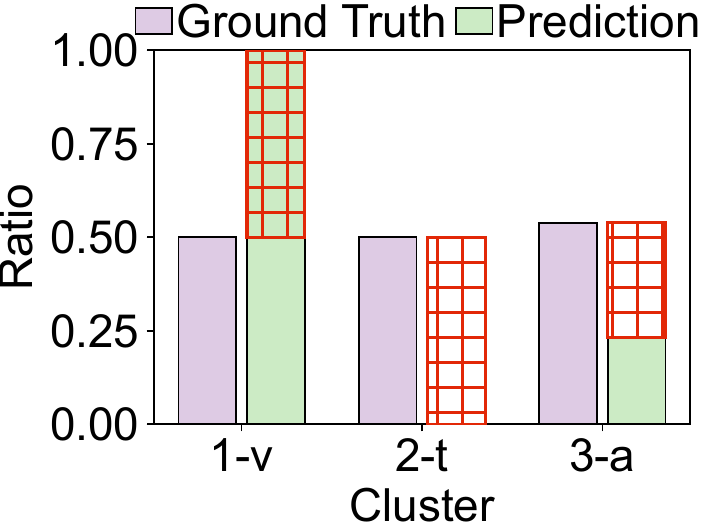}} 
    \subfloat[From MHY to MHB.]{\includegraphics[height=0.37\columnwidth]{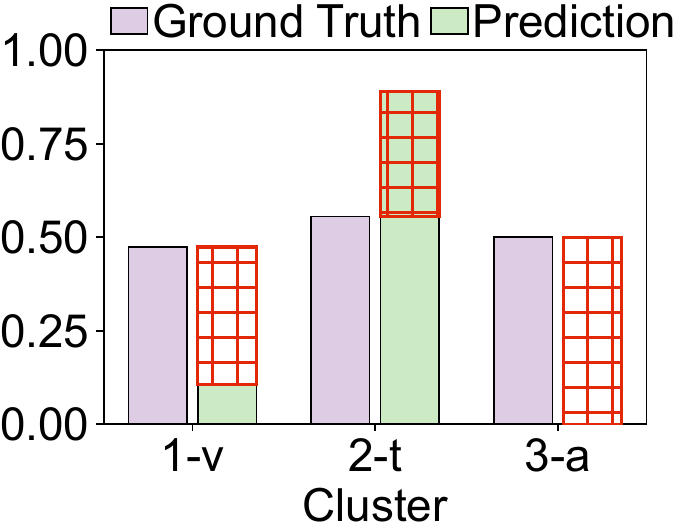}}
    
    \caption{
    Comparison of ground truth and predicted hateful video ratios (hateful vs. total videos) for one cluster per modality (1:visual, 2:textual, 3:audio) in SCAN. Grid regions indicate over- or under-estimation in predicted ratios relative to ground truth.}
    \label{fig:pre_epx1}
\end{figure}
\subsection{\M: Collapse-Avoidance Alignment}
\label{SCANNER}
While SCAN mitigates the influence of outliers, it may incur homogenizing outputs of target samples (i.e., all samples are predicted as hateful or non-hateful) when aligning modality-specific representations with their respective centroids.
As illustrated in~\Cref{fig:pre_epx1}, we report the distributions of both ground truth and predictions produced by SCAN framework on three selected clusters across two target domains (HMM and MHB), each cluster corresponding to a distinct modality (visual, textual, and audio). 
Although the ground truth distributions within each modality-specific cluster are balanced (with ratios close to 0.5), the corresponding prediction distributions are significantly imbalanced (with ratios close to 0 or 1), revealing that SCAN tends to assign the majority of target samples within a modality-specific cluster to a single, incorrect class, despite differences in their labels.
This phenomenon reflects a shortcut solution learned by the model during alignment, making the model prone to modality-level semantic collapse~\cite{hoang2024persistent}, where it generates overly homogeneous predictions around the centroids.

Therefore, to address this issue of the aforementioned semantic collapse, we incorporate an intra-cluster diversity loss into the SCAN framework, thereby developing the \textbf{S}ample-adaptive \textbf{C}entroid \textbf{A}lig\textbf{N}ment framework with i\textbf{N}tra-clust\textbf{ER} collapse avoidance (\textbf{\M}).
\M encourages high average prediction entropy at the cluster level, thereby promoting output diversity across samples within the same cluster.
We compute the average prediction probability $\bar{p}_{m,n}^t$ of the $n$-th cluster in modality $m$:
\begin{equation}
\label{eq:avg_prob}
\bar{p}_{m,n}^t = \frac{1}{|S_{m,n}|} \sum_{i \in S_{m,n}} \text{softmax}(o_{m,i}^t),
\end{equation}
where $o_{m,i}^t$ is the logit output for target sample $D_i^t$, and $S_{m,n}$ is the set of videos assigned to the $n$-th cluster for modality $m$, $|S_{m,n}|$ represents the number of videos in the $n$-th cluster.
We then formulate the diversity loss $\mathcal{L}_{\text{DIV}}$ as follows:
\begin{equation}
\label{eq:div_loss}
\mathcal{L}_{\text{DIV}} = \sum_{m \in \{v, t, a\}} \left( \frac{1}{k} \sum_{n=1}^{k} \left( \sum_{c=1}^{C_{cls}} \bar{p}_{(m,n),c}^t \cdot \log(\bar{p}_{(m,n),c}^t) \right) \right),
\end{equation}
where $C_{cls}$ is the number of classes and $\bar{p}_{(m,n),c}^t$ is the average prediction probability for class $c$ over the $n$-th cluster for modality $m$.
$\mathcal{L}_{\text{DIV}}$ encourages the average prediction distribution within each cluster to be as diverse as possible.
And results from~\Cref{fig:intra_diversity} demonstrate that, in contrast to SCAN, \M effectively mitigates the semantic collapse.

\subsection{Adaptation Objective}
During the test time, the source model parameters are updated using the following objective:
\begin{equation}
\label{eq:final_loss}
\mathcal{L} = \epsilon \cdot \mathcal{L}_{\text{EM}} + \lambda \cdot \mathcal{L}_{\text{SCAN}} + \alpha \cdot \mathcal{L}_{\text{DIV}},
\end{equation}
here $\epsilon$, $\lambda$, and $\alpha$ are hyperparameters. 
Following prior TTA methods~\cite{niu2022efficient}, we add an entropy minimization loss $\mathcal{L}_{\text{EM}}$ to encourage the source model to make more confident and clearer predictions on target samples during adaptation.
These objectives work in tandem to enhance the source model’s robustness to tackle the drastic semantic drift in evolving hateful videos, thereby improving generalization.

\section{Experiments}
\subsection{Experimental Setup}

\noindent\textbf{Datasets.}
As shown in~\Cref{tab:dataset}, we conduct experiments on three benchmarks in \task: HateMM (HMM)~\cite{das2023hatemm}, MultiHateClip-Youtube (MHY), and MultiHateClip-Bilibili (MHB)~\cite{wang2024multihateclip}.

\begin{table}[h]
\centering
\setlength{\tabcolsep}{2pt}
\begin{tabular}{lcccc}
\toprule
\textbf{Dataset} & \textbf{Total} & \textbf{Hateful(\%)} & \textbf{Platform} & \textbf{Language} \\
\midrule
HMM & 1,083 & 39.8 & BitChute & English \\
MHY & 1,000 & 33.8 & YouTube & English \\
MHB & 1,000 & 32.2 & Bilibili & Chinese \\
\bottomrule
\end{tabular}
\caption{Characteristics of three hateful video datasets.}
\label{tab:dataset}
\end{table}

\noindent\textbf{Setting of TTA.}
To evaluate the adaptability of our proposed \M under real-world distribution shifts in \task, we conduct cross-domain detection experiments.
Each dataset originates from a distinct platform (BitChute, YouTube, Bilibili), differs in language (English or Chinese), and exhibits unique class distributions, leading to substantial semantic drift, as summarized in~\Cref{tab:dataset}.
These inherent discrepancies allow each dataset to be treated as a distinct domain, simulating semantic drift caused by the evolution of hateful videos.
As a result, we define six evaluation groups, where each source-to-target domain adaptation setting is denoted as \textbf{Dataset A $\rightarrow$ Dataset B}.

\noindent\textbf{Baselines.} Baselines are categorized into two classes: 
(1) Traditional \task methods, which are pre-trained on the source domain dataset without adaptation, including HTMM~\cite{das2023hatemm}, MHCL~\cite{wang2024multihateclip}, Pro-Cap~\cite{cao2023pro}, and MoRE~\cite{lang2025biting}.
(2) TTA methods, which adapt the source model to the unlabeled target data without accessing the source data, including Source, Test-time normalization (Norm)~\cite{schneider2020improving}, Self-Training (ST)~\cite{liang2020we}, TENT~\cite{wang2021tent}, SAR~\cite{niu2023towards}, READ~\cite{yang2024test}, and SuMi~\cite{guo2025smoothing}.

\noindent\textbf{Evaluation Metrics.} Following prior works~\cite{lang2025biting}, we adopt two metrics to evaluate the model's performance: Classification Accuracy (ACC), Macro-F1 (M-F1).

\begin{table*}[h]
\setlength{\tabcolsep}{2.8pt}
\centering
\begin{tabular}{lcccccccccccccc}
\toprule
& \multicolumn{2}{c}{\textbf{MHY\textrightarrow  HMM}} & \multicolumn{2}{c}{\textbf{MHY\textrightarrow  MHB}} 
& \multicolumn{2}{c}{\textbf{HMM\textrightarrow MHB}} & \multicolumn{2}{c}{\textbf{HMM\textrightarrow MHY}} 
& \multicolumn{2}{c}{\textbf{MHB\textrightarrow HMM}} & \multicolumn{2}{c}{\textbf{MHB\textrightarrow MHY}}
& \multicolumn{2}{c}{\textbf{Avg.}} \\
\cmidrule(lr){2-3} \cmidrule(lr){4-5} \cmidrule(lr){6-7} \cmidrule(lr){8-9} \cmidrule(lr){10-11} \cmidrule(lr){12-13} \cmidrule(lr){14-15}
{Method} & \textbf{ACC} &  \textbf{M-F1} & \textbf{ACC} &  \textbf{M-F1} & \textbf{ACC} &  \textbf{M-F1} 
                & \textbf{ACC} &  \textbf{M-F1} & \textbf{ACC} &  \textbf{M-F1} & \textbf{ACC} &  \textbf{M-F1} & \textbf{ACC} &  \textbf{M-F1}\\
\midrule
HTMM & 59.46 & 57.33 & 56.00 & 50.55 & 58.30 & 50.50 & 64.80 & 47.50 & 45.71 & 45.56 & 53.00 & 52.11 & 56.21 & 50.59 \\
MHCL & 57.43 & 52.72 & 60.50 & 51.79 & 63.90 & 52.18 & 61.50 & 48.18 & 52.44 & 52.25 & 61.90 & 51.39 & 59.61 & 51.42 \\
Pro-Cap & 50.05 & 47.04 & 63.90 & 53.94 & 66.20 & 45.99 & 59.60 & 51.24 & 54.57 & 44.83 & 60.20 & 48.41 & 59.09 & 48.58 \\
MoRE & 55.03 & 54.71 & 65.70 & 44.39 & 37.10 & 32.87 & 50.20 & 50.13 & 51.25 & 36.87 & 61.30 & 43.34 & 53.43 & 43.72 \\
\midrule
Source & 62.94 & 56.81 & 64.00 & 57.38 & \underline{67.00} & 45.02 & 57.34 & 57.28 & 53.94 & 53.64 & 63.60 & \underline{59.00} & 61.47 & 54.86 \\
Norm & 63.62 & 57.14 & \underline{66.00} & 57.65 & 58.50 & 53.03 & 64.90 & 59.36 & 60.75 & 53.24 & 65.00 & 54.85 & 63.13 & 55.88 \\
ST & 60.66 & \underline{60.39} & \underline{66.00} & 54.17 & \textbf{67.60} & 40.33 & 65.00 & 40.97 & 52.72 & 52.72 & 66.10 & 39.83 & 63.01 & 48.07 \\
TENT & 63.06 & 56.19 & 65.80 & \underline{59.81} & 58.00 & 53.01 & 64.30 & 60.47 & 61.03 & 54.02 & \underline{66.30} & 58.07 & 63.08 & 56.93 \\
SAR & \underline{63.67} & 59.26 & 65.65 & 59.73 & 58.10 & \underline{53.15} & \underline{65.20} & \underline{61.28} & \underline{61.23} & \underline{55.70} & 66.10 & 57.91 & \underline{63.32} & \underline{57.84} \\
READ & 63.34 & 57.21 & 65.21 & 57.30 & 58.40 & 53.11 & 64.20 & 60.44 & 60.84 & 53.64 & 65.00 & 55.07 & 62.83 & 56.13 \\
SuMi & 63.34 & 56.85 & 65.60 & 56.82 & 57.00 & 52.23 & 63.70 & 59.75 & 60.94 & 53.63 & 66.20 & 56.05 & 62.80 & 55.89 \\
\midrule
\textbf{\M} & \textbf{68.14} & \textbf{64.63} & \textbf{68.30} & \textbf{60.57} & 60.20 & \textbf{56.49} & \textbf{66.70} & \textbf{62.90} & \textbf{62.69} & \textbf{58.59} & \textbf{66.89} & \textbf{60.10} & \textbf{65.49} & \textbf{60.55} \\
\bottomrule
\end{tabular}
\caption{
Performance comparisons between baselines and our \M across six scenarios derived from the MHY, MHB, and HMM datasets.
Bold and underlined values indicate the \textbf{best} and \underline{second-best} performances, respectively.}

\label{tab:main_result}
\end{table*}

\noindent\textbf{Implementation Details.} 
During the feature extraction phase, when the source and target domains are in the same language, we utilize the pre-trained CLIP~\cite{radford2021learning} for feature extraction.
In other cases, we employ multilingual Sentence-BERT~\cite{reimers-2019-sentence-bert} to ensure effective cross-lingual representation extraction.
Following previous works~\cite{ma2024improved, guo2025bridging}, we use transformer~\cite{vaswani2017attention} as the source models. 
During the online test-time adaptation phase, we use AdamW optimizer, with an initial learning rate of $1.0\times10^{-3}$, weight decay of $5\times10^{-4}$ and batch size of 128. 
The overall process is conducted within a single epoch.
For learnable parameters, we update affine parameters in normalization layers and the linear layer in modal-specific encoders by following the prior works~\cite{niu2023towards, yang2024test}.

\subsection{Main Performance}
\label{Main_Performance}

We report the performance in~\Cref{tab:main_result}.
From the results, we have the following observations:

\begin{table}[t]
    \centering
    \setlength{\tabcolsep}{3pt}
    \resizebox{\linewidth}{!}{
    \begin{tabular}{ccccccc}
    \toprule
     \multicolumn{3}{c}{\textbf{Framework}} & \multicolumn{2}{c}{\textbf{MHY\textrightarrow HMM}} & \multicolumn{2}{c}{\textbf{MHY\textrightarrow MHB}}\\
     \cmidrule(lr){1-3} \cmidrule(lr){4-5} \cmidrule(lr){6-7} 
     \textbf{CAN} & \textbf{SACN} & \textbf{\M} & \textbf{Acc} & \textbf{M-F1} &\textbf{Acc} & \textbf{M-F1} \\ \midrule
     \ding{55} & \ding{55} & \ding{55} & 62.94 & 56.81 & 64.00 & 57.38 \\  
     \ding{51} & \ding{55} & \ding{55} & 66.45 & 62.17 & 65.40 & 58.57 \\  
     \ding{55} & \ding{51} & \ding{55} & 67.74 & 64.02 & 65.72 & 58.95 \\  

     \ding{55} & \ding{55} & \ding{51} 
     & \textbf{68.14} & \textbf{64.63} & \textbf{68.30} & \textbf{60.57} \\ 
     \bottomrule
    \end{tabular}}
    \caption{Ablation studies comparing different variants of the framework. The best results are in black \textbf{bold}.}
    \label{tab:ablation1}
\end{table}

\noindent\textbf{O1: Superiority of TTA methods over traditional \task approaches.}
Traditional \task methods typically assume that the training (source) and inference (target) data share the same distributions, i.e., the assumption of distribution consistency.
Therefore, these models fail drastically when faced with large distributional drift caused by variations in hateful manifestations.
In contrast, TTA methods bridge the domain gap by employing entropy minimization (EM) to address mild distribution shifts. 
Therefore, they outperform traditional \task methods under such conditions.

\noindent\textbf{O2: SCANNER outperforms baseline methods in overall performance.}
Existing TTA methods are designed to cope with mild and structured distribution shifts. 
However, in the context of \task, where target-domain hateful videos appear in highly diverse and irregular surface manifestations, 
resulting in a huge semantic drift that makes TTA methods fall short.
Our proposed \M addresses these challenges by explicitly aligning target videos to stable centroids. 
This reduces the source model's sensitivity to superficial variations in manifestations and effectively bridges the distribution gap in source and target domains.
In addition, traditional methods tend to exhibit bias under severe class imbalance by over-fitting to the majority class, whereas our \M explicitly encourages correct prediction of minority-class samples through centroid-guided alignment, demonstrating strong robustness against over-fitting.

\subsection{Ablation Study}

To further understand the effectiveness of each progressively upgraded variant of our frameworks, we conduct ablation studies. 
The performance of each framework is summarized in~\Cref{tab:ablation1}. 
The base framework \textbf{CAN} demonstrates improved detection performance over direct inference using the source model. 
This is attributed to its centroid-guided alignment, which aligns targets to corresponding invariant hateful centroids, thereby reducing sensitivity to diverse hateful manifestations.
Building on this, \textbf{SCAN} incorporates a sample-adaptive alignment strategy that mitigates the adverse impact of low-similarity samples during alignment, thereby improving detection performance and stabilizing optimization.
To address intra-cluster homogenization --- where representations collapse toward the centroid --- an intra-cluster diversity loss is introduced into the SCAN framework, resulting in the final framework \textbf{SCANNER}. 
This design preserves representational diversity within clusters and yields superior detection performance.
\begin{figure}[t]
    \centering
    \subfloat[From MHY to HMM.]{\includegraphics[height=0.35\columnwidth]{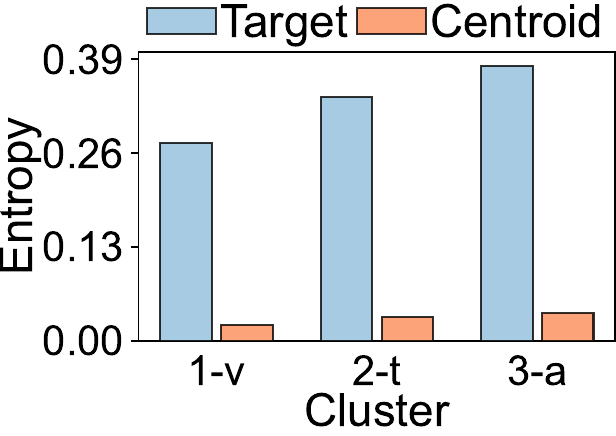}} 
    \subfloat[From MHY to MHB.]{\includegraphics[height=0.35\columnwidth]{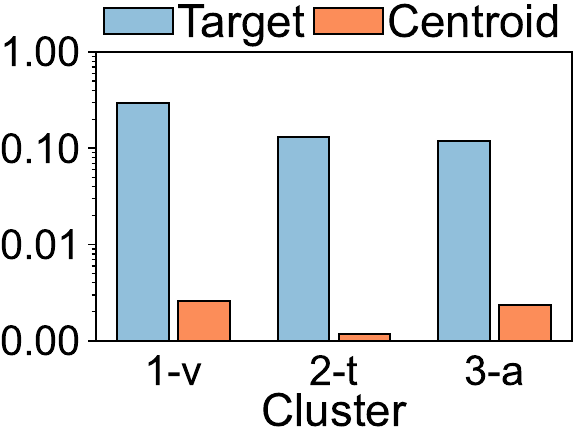}}
    \caption{Entropy comparison between modality-specific cluster (1:visual, 2:textual, 3:audio) centroids and the average entropy of target videos within corresponding cluster across two target domains.}
    \label{fig:Centroids}
\end{figure}

\subsection{Centroids Stability Evaluation}

To assess the rationality of the centroid-guided alignment in \M, we select one cluster from each modality. For each modality-specific cluster, we compute and compare the prediction entropy of its centroid with the average prediction entropy of the associated target videos.
\Cref{fig:Centroids} reveals that the source model yields high entropy for target samples, intuitively reflecting a notable domain gap. 
This high-entropy phenomenon indicates that the source model struggles with confident predictions for target videos due to novel and diverse manifestations. 
However, the corresponding modality-specific centroids formed by clustering these target samples exhibit significantly lower prediction entropy. 
These low-entropy centroids effectively capture the underlying core semantics and intents shared across domains, acting as the cores that bridge the target domain back to the source.
Hence, our framework guides uncertain target samples toward these stable centroids for more robust adaptation.

\subsection{Intra-cluster Diversity Evaluation} 

\begin{figure}[t]
    \centering
    \subfloat[From MHY to HMM.]{\includegraphics[height=0.37\columnwidth]{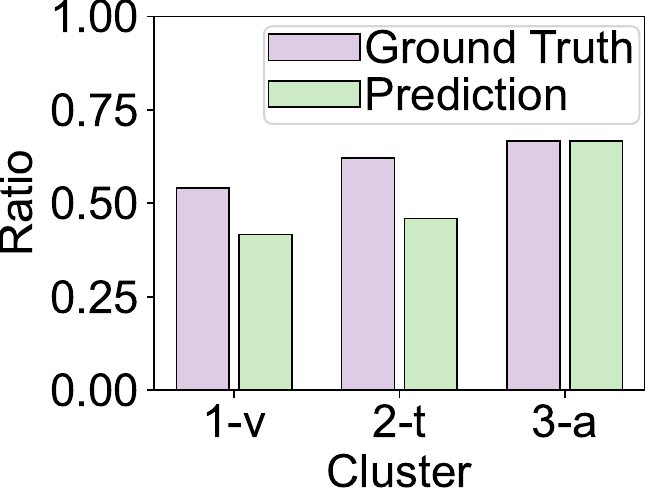}} 
    \subfloat[From MHY to MHB.]
    {\includegraphics[height=0.37\columnwidth]{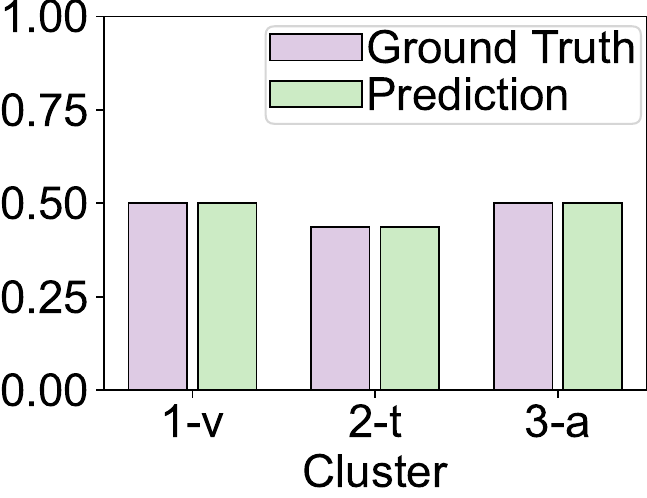}}
    \caption{Ground truth vs. predicted hateful ratios for one cluster per modality (1:visual, 2:textual, 3:audio). Smaller bar gaps indicate better intra-cluster prediction consistency.}
    \label{fig:intra_diversity}
\end{figure}
To evaluate the effectiveness of our intra-cluster diversity regularization, we compare the ground-truth and predicted hateful ratios within one cluster randomly selected from each modality across two target domains, as shown in~\Cref{fig:intra_diversity}. 
The ground-truth class distributions within these clusters are relatively balanced, and our framework yields similarly balanced predictions, with class ratios consistently near 0.5. 
These results indicate that our method effectively mitigates the semantic collapse issue (cf.~\Cref{fig:pre_epx1}) by preserving class diversity and preventing the source model from assigning all target samples to a single incorrect class.

\subsection{Domain Distribution Visualization} 
\begin{figure}[ht]
    \centering
{\includegraphics[width=0.45\columnwidth]{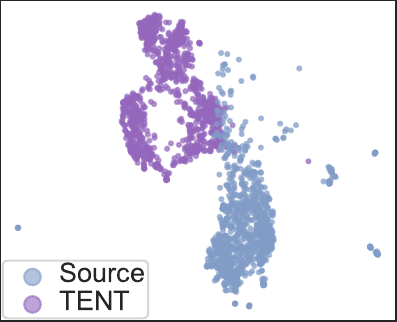}} 
{\includegraphics[width=0.45\columnwidth]{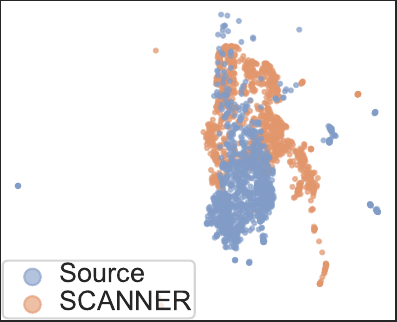}}
\caption{UMAP visualizations of embedding distributions for the source (MHY) and adapted target (HMM) domains on TENT and \M. }
\label{fig:umap}

\end{figure}

To intuitively compare the capability of \M and TENT in reducing domain discrepancy between source and target domains, we present UMAP visualizations~\cite{McInnes2018} of embedding distributions from the source (MHY) and adapted target (HMM) domains, as shown in~\Cref{fig:umap}. 
Final representations are obtained by averaging outputs of modality-specific encoders. 
While TENT adapts via EM, the resulting source and target embeddings remain clearly separated, suggesting that implicit alignment fails to overcome substantial semantic shifts. 
In contrast, \M achieves stronger alignment with considerable overlap between domains. 
This improvement stems from \M’s use of invariant hateful cores as domain-bridging anchors that explicitly guide target samples toward source-aligned representations, effectively mitigating the influence of surface-level variations in evolving hateful content.

\subsection{Case Study on Ambiguous Hateful Videos} 
\begin{figure}[t]
    \centering
    {\includegraphics[width=0.98\columnwidth]{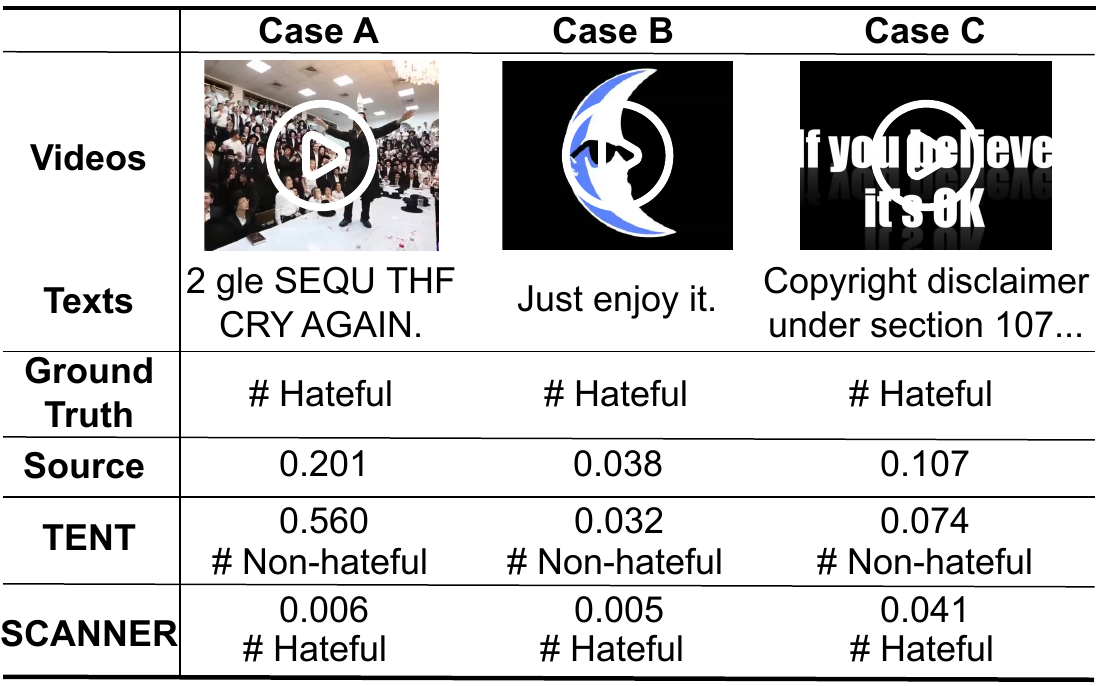}}
    \caption{Case study of \M's adaptability on three randomly sampled hateful videos from the target domains. The numerical result is the prediction entropy.}       
    \label{fig:case}

\end{figure}

We examine the adaptability of \M on three representative target hateful videos, as illustrated in~\Cref{fig:case}. 
In all cases, both visual and textual content lack explicit hateful cues; the presented semantics are abstract and implicitly expressed, making it difficult to discern any hateful intent. 
Compared to the initial entropy produced by the source model, TENT exhibits limited adaptation capability, often leading to incorrect predictions and even performance degradation. 
In the particularly challenging Case A, TENT demonstrates high uncertainty toward subtly expressed hateful content, ultimately resulting in misclassifications. 
In contrast, \M explicitly aligns target videos with the source domain via invariant hateful cores, yielding the lowest entropy values and correctly classifying all samples.

\section{Conclusion}

In this paper, we propose \M, the first TTA framework tailored for \task. 
We adopt a progressive expansion scheme, starting with a basic framework, dubbed CAN, which initially guides target samples toward the semantic centroids representing the invariant hateful cores. 
To further mitigate the impact of video samples distant from the semantic centroids, we extend CAN to SCAN by introducing sample-level adaptive weighting based on sample-to-centroid similarity.
Finally, we incorporate an intra-cluster diversity loss into SCAN to avoid representation homogenization within cluster, thus forming the final framework, \M.
Extensive experiments on three benchmarks show that \M outperforms all baselines in \task. 

\section*{Acknowledgments}
This work was supported by National Natural Science Foundation of China (Grant No. 62572097, No. 62176043, and No.U22A2097).

\bibliography{aaai2026}
\appendix

\end{document}